# Linear vs Nonlinear Extreme Learning Machine for Spectral-Spatial Classification of Hyperspectral Image

**Faxian Cao [1], Zhijing Yang [1,*], Jinchang Ren [2], Mengying Jiang [1] and Wing-Kuen Ling [1]**

[1] School of Information Engineering, Guangdong University of Technology, Guangzhou, 510006, China; faxiancao@foxmail.com; yzhj@gdut.edu.cn; 346047000@qq.com; yongquanling@gdut.edu.cn

[2] Department of Electronic and Electrical Engineering, University of Strathclyde, Glasgow, G1 1XW, UK; jinchang.ren@strath.ac.uk

* Correspondence: yzhj@gdut.edu.cn; Tel.: +86-20-39322438

**Abstract:** As a new machine learning approach, extreme learning machine (ELM) has received wide attentions due to its good performances. However, when directly applied to the hyperspectral image (HSI) classification, the recognition rate is too low. This is because ELM does not use the spatial information which is very important for HSI classification. In view of this, this paper proposes a new framework for spectral-spatial classification of HSI by combining ELM with loopy belief propagation (LBP). The original ELM is linear, and the nonlinear ELMs (or Kernel ELMs) are the improvement of linear ELM (LELM). However, based on lots of experiments and analysis, we found out that the LELM is a better choice than nonlinear ELM for spectral-spatial classification of HSI. Furthermore, we exploit the marginal probability distribution that uses the whole information in the HSI and learn such distribution using the LBP. The proposed method not only maintain the fast speed of ELM, but also greatly improves the accuracy of classification. The experimental results in the well-known HSI data sets, Indian Pines and Pavia University, demonstrate the good performances of the proposed method.

---

## 1. Introduction

Classification is the basic research and an important mean of obtaining information from hyperspectral images (HSI). The main goal of HSI classification is to divide each pixel of an image into different classes according to the spectral information and the spatial information. Since each pixel of HSI has many spectral features, it is difficult to classify HSI with limited samples and high spectral resolution, which is a challenging problem for HSI classification. There are some typical algorithms for HSI images classification, such as support vector machine (SVM) [2], sparse multinomial logistic regression (SMLR) [3], and so on. Many techniques have been proposed for feature extraction and dimensionality reduction [4,5], such as singular spectrum analysis (SSA) [6-9], principal component analysis (PCA) [10,11] and spectral-spatial classification methods [12] and so on. However, there are still many challenges for HSI classification, for example, the data structure of each pixel in the HSI data is very complex, and each pixel of HSI data sets has very large dimensions. Therefore, it is very difficult to not only reduce the consuming time for classification, but also achieve high accuracy for classification with little training samples.

As a new machine learning approach that has single-hidden layer feedforward neural network, ELM has received wide attentions due to its good performances. It has been proved to be a promising algorithm in pattern recognition fields [13-17]. Compared to support vector machine and other state-of-the-art algorithms, ELM has the following advantages [17]: very simple structure and higher generalization, high computational efficiency without tuning additional parameters. The original ELM is a linear operation, so we call it linear ELM (LELM). Although it has the above advantages, the classification accuracy is not very high when applied to hyperspectral images. Kernel ELM (KELM) [18] and sparse ELM [19] are the improvements of LELM and achieve better classification results for pattern recognition. The accuracy of KELM is improved but still not high enough when applied to the classification of HSI. So it is a critical problem that not only maintaining the property of fast speed, but also improving the classification accuracy for HSI classification with ELM. The main reason that LELM and KELM cannot achieve high accuracy of classification is that they just make use of spectral information of HSI, without the spatial information of HSI. The spatial information, which reflects the local property of HSI data sets, is very important for classification.

To improve the performance of ELM for HSI classification, Loopy belief propagation (LBP) algorithm is used here [20,21]. It is a conditional probability model, which can be considered as a generalization of the Markov chain and can effectively describe the correlation of all the nodes/pixels in the field. It is based on the Markov random field (MRF) which assumes that the neighboring pixels likely belong to the same class [22-24]. The principle of LBP for classification is to calculate the marginal probability based on the characteristics of the samples. The KELM is the improvement of ELM, and it has been combined with MRF for spectral and spatial classification of HSI [25]. It can improve the recognition result of KELM to some extent. However, based on lots of experiments and analysis, we found out that the linear ELM is a better choice than KELM for spectral-spatial classification of HSI.

LELM is a type of linear operation, so its final mapping results will not change the characteristics of pixels in HSI. Nevertheless, KELM is a type of nonlinear operation, so called NLELM, and its final mapping results will disturb the features of pixels in the same class. If we use the output of NLELM as the input of MRF or LBP, the structure of NLELM will seriously disturb the original information of HSI. Then it cannot fully utilize the spectral information and spatial information of HSI and will cause the classification accuracy relative low. For example, the NLELM and MRF are combined for classification of HSI in [25], called NLELM-MRF. NLELM disturbs the features of pixels in the same class and causes the classification accuracy relative low. The kernel form is a type of nonlinear operation, so it will disturb the features of pixels in HSI, and cause the classification results relative low. In summary, we should not disturb the features of pixels in HSI before using spatial information extracted by LBP for improving the classification accuracies of HSI. Hereby, LELM is used here with LBP for spectral-spatial classification of HSI to achieve high classification accuracy.

As mentioned above, LBP algorithm is based on the MRF. The LBP uses the information of the node and the node to transmit information to update the current MRF marking state [3]. It is a kind of approximate calculation based on MRF. This algorithm is an iterative method, which can solve the problem of probabilistic inference in probabilistic graphical models. After many iterations of probability, the belief of all the nodes is no longer changed. Then the LBP algorithm can converge to its optimal solution. Since the pixels of HSI that need to be classified are just a part of HSI, it means that not all the pixels in HSI need to be classified. If we use LBP to classify HSI directly, it may cause ill-posed problems. In view of this, we make some improvement of LBP for HSI classification. The pixels of background of HSI are ignored in the process of LBP. The proposed framework will fully make use of the spectral and spatial information by ELM to improve the classification accuracy dramatically. Experiment results demonstrate the better performance compared with other state -of-the-art methods at the same situation.

The remaining of this paper is divided into the following sections: Section 2 describes the experimental data and the detail of the proposed method. Section 3 shows the extensive experimental results and analysis. Conclusions are summarized in Section 4.

## 2. Materials and Methods

In this section, we first introduce the experimental data sets, then we elaborate the proposed method based on LELM and LBP.

### *2.1. HSI Data Set*

The experimental data sets include two well-known HSI datasets, which are detailed below.

(1) *Indian Pines:* The Indian Pines HSI data set was the urban image collected in June 1992 by the AVIRIS sensors over the Indian Pines region. The data set has 145×145 pixels which each has 200 spectral bands after removing 20 water absorption bands ranging from 0.2 to 2.4 μm. There are totally 16 classes.

(2) *Pavia University:* The Pavia university HSI data set was acquired in 2001 by the Reflective Optics System Imaging Spectrometer, flown over city of Pavia Italy. The sensor collects HSI data set in 115 spectral bands ranging from 0.43 to 0.86 μm with a spatial resolution of 1.3m/pixel. 103 bands were selected for experiment after removing 12 noisiest bands. The image scene contains 610×340 pixels and there are totally 9 classes.

### *2.2. Normalization*

Let $X \equiv (X_1, X_2, \dots, X_N) \in R^{N\times d}$ be HSI data, which has N samples and each sample has d features. Normalization is a preprocessing process and has a great influence on the subsequent classification of data. Based on lots of experiments, we choose the stable normalization method as follows:

$$x_{ij} = X_{ij}/\max(X) \quad (1)$$

where $X_{ij}$ is any pixel value of the HSI data, max() is the largest value of all the data in the HSI.

*2.3. Linear ELM*

For convenient, let $x \equiv (x_1, x_2, \dots, x_N) \in R^{N\times d}$ be the HSI data after normalization, $y \equiv (y_1, y_2, \dots, y_N) \in R^{N\times M}$ denotes the class labels. As a new learning algorithm, ELM [17] is a single layer feedforward neural network, which can be modeled as:

$$\sum_{j=1}^{L} \beta_j G(w_j^T x_i + b_j) = y_i \quad (2)$$

where $w_i = (w_{i1}, w_{i2}, \dots, w_{iL})^T$ is the weight vector connecting the input layer with hidden layer of *i*-th sample; $b_i$ is the bias connecting input layer with hidden layer of *i*-th sample and $\beta_j$ is the output weight vector of *i*-th sample; *T* is the transpose operation and *g()* is the activation function of the hidden layer. The main steps of classification with ELM are as follows:

**Step1:** Assign random input $w_i$ and bias $b_i$, $i = 1, 2, \dots, N$ for the input layer.

**Step2:** Calculate the output matrix of hidden layer *G* as:

$$G(w_1, w_2, \dots, w_N; x_1, x_2, \dots, x_N; b_1, b_2, \dots, b_N) = \begin{bmatrix} g_{11}(w_{11}x_{11} + b_{11}) & \dots & g_{1L}(w_{1L}x_{1L} + b_{1L}) \\ \dots & \dots & \dots \\ g_{N1}(w_{N1}x_{N1} + b_{N1}) & \dots & g_{NL}(w_{NL}x_{NL} + b_{NL}) \end{bmatrix} \quad (3)$$

**Step3:** Calculate the output matrix **β:**

$$\beta = G^{\dagger} y \quad (4)$$

where $\beta = [\beta_1, \dots, \beta_L]_{L\times M}^T$ and $\dagger$ is the Moore-Penrose generalized inverse of hidden layer matrix.

**Step4:** The result of the final classification of ELM can be expressed by the following equation:

$$f(x) = G * \beta \quad (5)$$

The execution time of ELM can be greatly reduced because the input weight and bias of ELM are randomly generated, so the output weight can be directly computed as $\beta = G^{\dagger} * y$. Any piecewise continual function can be used as the hidden layer activation function. Obviously, ELM is a lineal operation.

*2.4. Nonlinear ELM*

The classification problem for NLELM [22] can be formulated as:

$$Minimize: L_{NLELM} = \frac{1}{2} \parallel \beta \parallel_F^2 + C\frac{1}{2}\sum_{i=1}^{N} \parallel \varepsilon_i \parallel_2^2$$

$$subject\ to: \quad h(x_i)\beta = t_i^T - \varepsilon_i^T,\ i=1,\ \dots,\ N \quad (6)$$

where $\varepsilon_i = [\varepsilon_{i,1}, \dots, \varepsilon_{i,M}]$ is the error vector of the M output nodes relative to the sample $x_i$. $h(x_i)$ is the output of *i*-th sample between hidden layer and input layer. Based on the KKT theorem, equation (6) is equivalent to solve the following dual optimization problem:

$$L_{NLELM} = \frac{1}{2} \parallel \beta \parallel_F^2 + C\frac{1}{2}\sum_{i=1}^{N} \parallel \varepsilon_i \parallel_2^2 - \sum_{i=1}^{N}\sum_{j=1}^{M} \alpha_{i,j}\,(h(x_i)\beta_j - t_{i,j}^T + \varepsilon_{i,j}^T) \quad (7)$$

where $\beta_j$ is the vector of weight between hidden layer and output layer. $\alpha_{i,j}$ is the Lagrange multiplier. Based on the KKT theorem, we can conclude that:

$$\frac{\partial L_{NLELM}}{\partial \beta_j} = 0 \rightarrow \beta = H^T \alpha \quad (8)$$

$$\frac{\partial L_{NLELM}}{\partial \varepsilon_i} = 0 \rightarrow \alpha_i = C\varepsilon_i \quad (9)$$

$$\frac{\partial L_{NLELM}}{\partial \alpha_i} = 0 \rightarrow h(x_i)\beta_j - t_i^T + \varepsilon_i^T \quad (10)$$

where $i$=1,…,N,$\alpha_i = [\alpha_{i,1}, \alpha_{i,2}, \dots\ , \alpha_{i,M}]^T$ and $\alpha = [\alpha_1, \alpha_2, \dots\ , \alpha_N]^T$. Now the output weight $\beta$ can be formulated as:

$$\beta = (\frac{I}{C} + H^T H)^{-1} H^T y. \quad (11)$$

The hidden neurons are unknown. Any kernel satisfying the Mercer's conditions can be used:

$$\mathbf{\Omega}_{KELM} = HH^T: \Omega_{KELM}(x_i, x_j)\mathrm{h}(x_i)\mathrm{h}(x_j)^T = K(x_i, x_j) \quad (12)$$

In general, the Gaussian kernel is chosen:

$$K_{NLELM}(x_i, x_j) = \exp(-\frac{\|x_i - x_j\|^2}{2 * \sigma_{NLELM}}) \quad (13)$$

Then the NLELM can be constructed using the kernel function.

There is no doubt that NLELM can achieve higher classification accuracy than LELM if we just consider the spectral information. As mentioned above, LELM is a linear operation for classification, and NLELM is nonlinear operations. The nonlinear operation is better than the linear operation in some aspects. However, the nonlinear operation will disturb the original features of the HSI data. If the subsequent classification needs to use the spatial information for classification, it will cause the classification accuracy relative low. So we will choose the LELM with LBP for spectral-spatial classification of HSI.

### *2.5. Using Spatial Information to Improve the Classification Accuracy Based on LBP*

To further extract the spatial information, the output of LELM is used as the input of LBP. The posterior density $\mathrm{p}(y/x)$ is obtained according to the feature $x$ that it is the output of LELM. We adopt the discriminative random field (DRF) [26] as:

$$\mathrm{P}(y/x) = \frac{1}{Z(x)}\exp(\sum \log p(y_i/x_i) + \sum \log p(y_i, y_j)) \quad (14)$$

where $Z(x)$ is the partition function. The term $\log p(y_i/x_i)$ is the association potential that model the likelihood of label $y_i$ given the feature $x_i$, and $\log p(y_i, y_j)$ is the interaction potential.

We adopt an isotropic MLL prior to the model image of class label $\mathrm{y}$ in order to use the spatial information of HSI. This prior belongs to the MRF class and encourages piecewise smooth segmentations. It tends to produce solutions that the adjacent pixels are likely to belong to the same class [3]. The MLL prior has been widely used in image segmentation problems [27-30] and is a generalization of the Ising model [31-33]. It can be formulated as:

$$\mathrm{p(y)} = \frac{1}{Z} exp^{\mu \sum \delta(\mathrm{y_i}, \mathrm{y_j})} \quad (15)$$

where $\mu$ is a tunable parameter controlling the degree of smoothness, Z is a normalization constant for the density, $\delta(\mathrm{y})$ is the unit impulse function. The pairwise interaction term $\delta(\mathrm{y_i}, \mathrm{y_j})$ assigns high probability to the neighborhood labels.

Maximum a posterior (MAP) estimate will minimize the Bayesian risk associated to the zero-one loss function [3]. The MAP estimate of $\mathrm{y}$ can be given by:

$$y^{\wedge} = \arg\min_{y} \sum -\log(\mathrm{y_i}/x_i) - \mu \sum \delta(\mathrm{y_i} - \mathrm{y_j}) \quad (16)$$

This is a combinatorial optimization problem having pairwise interaction terms. An alternative MAP solution is the MAP marginal (MAM) solution, which minimizes the Bayesian risk associated to the zero-one loss function. The MAM estimation of label $\mathrm{y_i}$ can be formulated as:

$$y_i^{\wedge} = \arg\max_{y_i} q(y_i/x) \quad (17)$$

where $q(y_i/x)$ is the marginal density of $\mathrm{p}(y/x)$ respect to $y_i$. The computation of marginal density of $\mathrm{p}(y/x)$ in (14) is difficult [3]. Since the LBP is an efficient approach to estimate Bayesian beliefs [20] in graphical model, we will use LBP to estimate the MAM solution and let the output of LELM $y_{LELM}^*$ be the input of LBP.

Figure 1 is a graphical example of MRF, where each node represents a random variable or a hidden node, and the class label $y_i$ here is associated with each input feature $x_i$. In the graphical example of MRF, $\psi_{ij}(y_i, y_j) = p(y_i, y_j)$ denotes the interaction potential that penalizes the dissimilar pair of neighboring label. $\varphi_i(y_i, x_i) = p(y_i/x_i)$ stands for the association potential of label $y_i$ respect to evidence. Suppose we observe some information about $x_i$. Each node has the state value $y_i$, and the observation value $x_i$. $\varphi_i(y_i, x_i)$ reflects the existence of statistical dependence. $\psi_{ij}(y_i, y_j)$ is the potential energy between adjacent neighbor nodes, and reflects the compatibility between the node variables $y_i$and $y_j$.

Figure 2 provides a graphical example of an undirected network. Since LBP is an iterative algorithm, at the *t*-th iteration, the message sent from node *i* to its neighbor node $j \in N(i)$ can be given by the following equation:

$$m_{ij}^t(y_j) = \frac{1}{Z}\sum_{y_i} \psi\,(y_i, y_j)\varphi(y_i, x_i)\prod_{k\in N(i)\backslash\{j\}} m_{ki}^{t-1}(y_i) \tag{18}$$

where *Z* is a normalization constant.

Assume that $b_i^t(y_i)$ is the belief of node *i* at the *t*-th iteration, it can be represented by the following equation:

$$b_i^t(y_i = k) = q(y_i = k/x) = \varphi(y_i = k)\prod_{j\in N(i)} m_{ji}^t(y_i = k) \tag{19}$$

Finally, we can estimate the final solution using maximize of the posterior marginal for node *i*:

$$y_i^{\wedge} = \arg\max_{y_i} q(y_i/x) = \arg\max_{y_i} b_i^t(y_i) \tag{20}$$

As we know, not all the pixels, but only part of the HSI needs to be classified. For instance, the size of HSI data set of Indian Pines is 145×145× 200, so the size of ground-truth is 145×145. But only 10366 out of 21025 pixels need to be classified. It may cause ill-posed problems if we use LBP directly with all the pixels. In view of this, we do some improvement of LBP (ILBP) in order to solve this problem, where we discard the pixel that belongs to the background, i.e. we just consider the pixels that need to be classified. The proposed method is summarized in Algorithm 1.

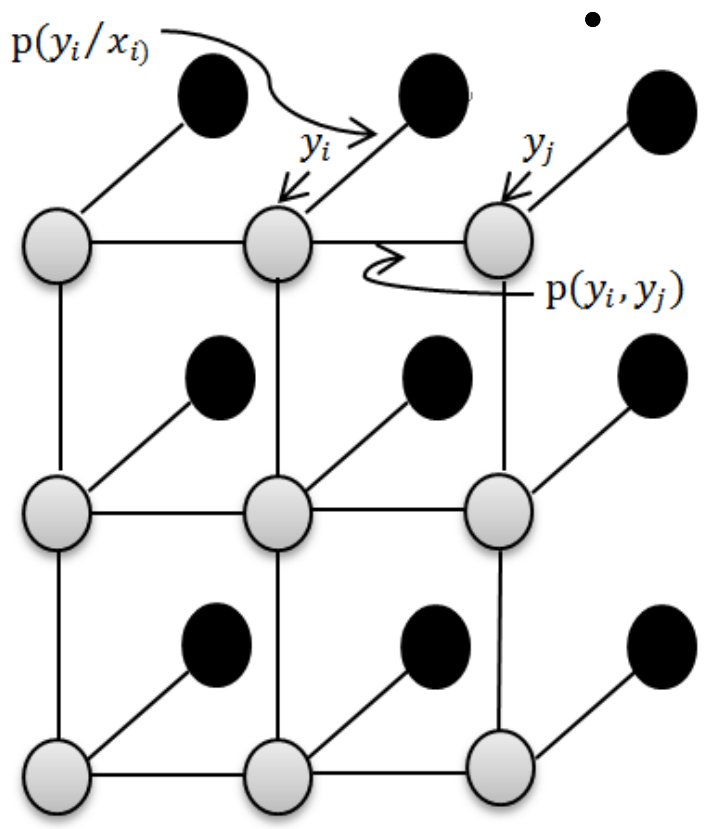

Figure 1. Graph example of MRF.

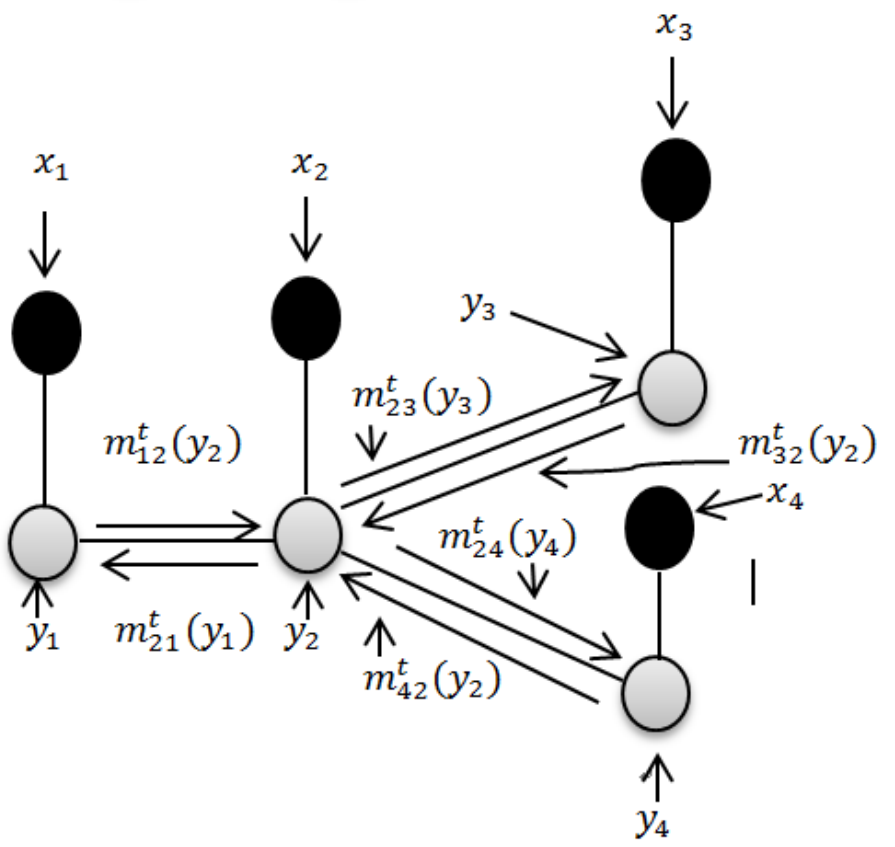

Figure 2. Message passing of LBP at *t*-th iteration.

---

**Algorithm 1:** Spectral-Spatial Classification for HSI Based on LELM and ILBP

---

**Input X:** the HSI image; X1: training samples; X2: test samples; Y1: The desired output of training sample; Y2:The desired output of test sample; *L*: number of hidden node of ELM; *g()*: activation function of hidden layer of ELM.

(1) **Normalization:** Let $X1^* = X1/\max(X)$, $X2^* = X2/\max(X)$.

(2) **LELM training:**

Step 1: Randomly generate the input weights, $w_i$, and bias, $b_i$.

Step 2: Calculate the hidden layer of output matrix:

$$G1 = g(w_i^T * X_1^* + b_i)$$

Step3: Calculate the output weight:

$$\beta = G^{\dagger} * Y1$$

**Output of LELM:** Calculate the hidden layer matrix of the test samples: $G2 = g(w_i^T * X2^* + b_i)$.

Get the output result of LELM: $Y_{ELM} = G2 * \beta$.

(3) **Spatial Classification by ILBP:**

Step1: Find the index of adjacent pixels of training samples and test samples and eliminate the pixels of background.

Step2: Calculate the marginal of MPA as follows:

For t=1: time of iterations

For j=1: number of pixels

If j~=test samples

Don't calculate the marginal of MAM.

Else

Calculate the marginal of MAM:

$$m_{ij}^t(y_j) = \frac{1}{Z}\sum_{y_i} \psi(y_i, y_j)\phi(y_i, x_i) \prod_{k\in N(i)\backslash\{j\}} m_{ki}^{t-1}(y_i)$$

Then the belief of node *i* at the *t*-th iteration can be represented as:

$$b_i^t(y_i = k) = q(y_i = k/x) = \phi(y_i = k) \prod_{j\in N(i)} m_{ji}^t(y_i = k)$$

End

End

The final solution for node i can be obtained by maximizing the posterior marginal:

$$y_i^{\wedge} = \arg\max_{y_i} q(y_i/x) = \arg\max_{y_i} b_i^t(y_i).$$

## 3. Results and discussions

In this section, the proposed method will be evaluated and relevant results are summarized and discussed in details. The experimental datasets include two well-known HSI datasets, i.e. Indian Pines and Pavia University. The number of training and test samples of each class is shown in Table 1.

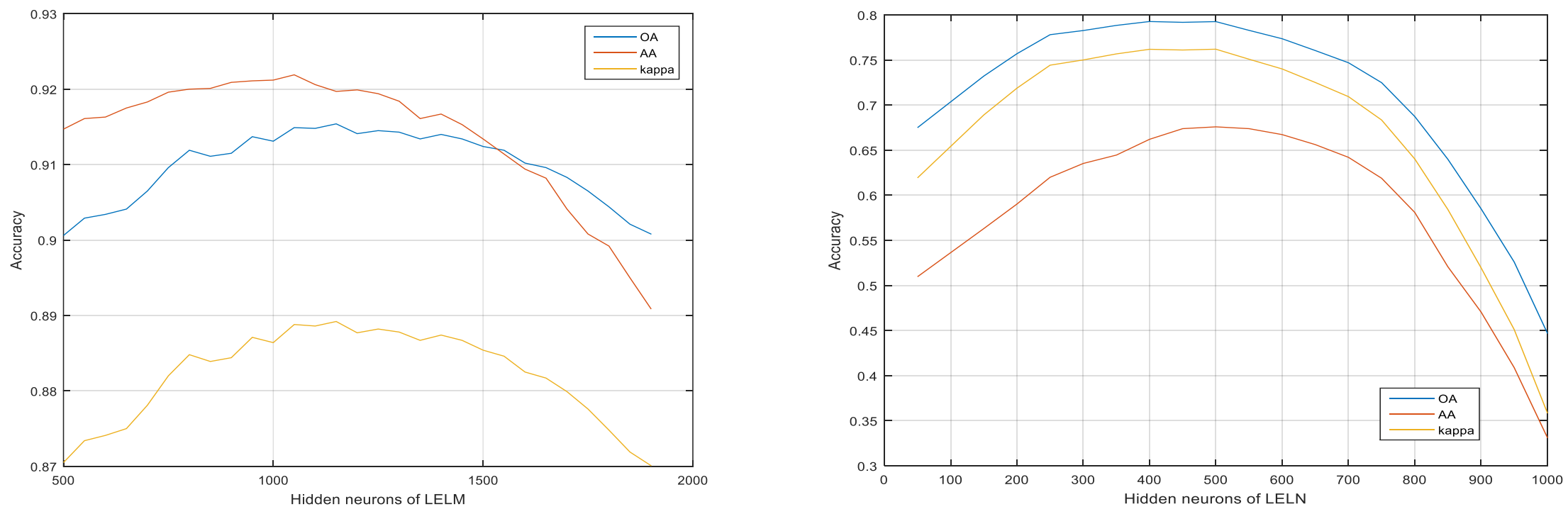

(**a**) (**b**)

**Figure 3.** The impact of hidden neurons of ELM in the datasets: (a) Indian Pines; (b) Pavia University.

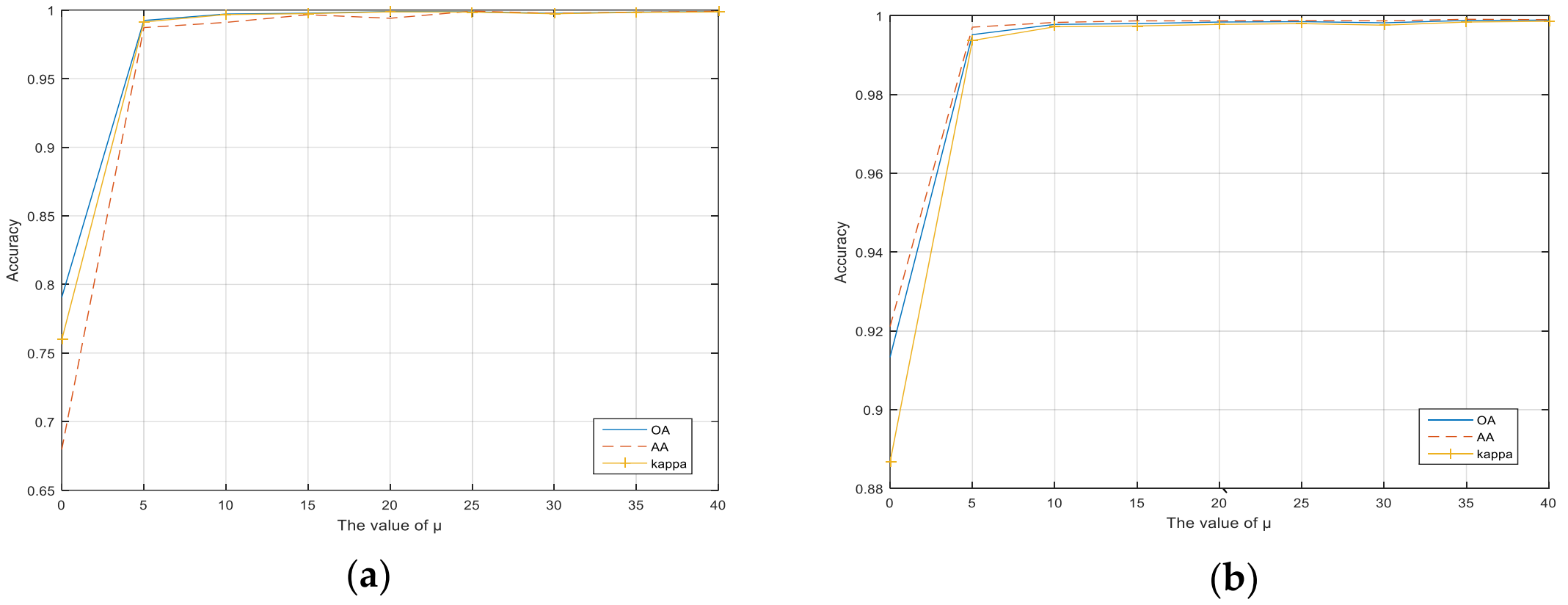

**Figure 4.** The impact of sparseness parameters μ in the datasets: (a) Indian Pines; (b) Pavia University.

### 3.1. Parameter Settings

All the experimental results are assessed by the overall accuracier (OA), averge accuracies (AA) and kappa stastistic (k) [31]. In order to avoid the influences induced by the selection of training samples, ten independent Monte Karlo runs are perfomed and OA, AA, k are all averged by ten runs.

In order to compare the performance of the proposed method with other classifiers and avoid the impact induced by the parameter setting, we show the parameter settings in the experiments. The parameters of SMLR and KSMLR are the same as [26]. (It should be noting that the SMLR and KSMLR are SMLR and KSMLR via variable splitting and augmented Lagrangian (LORSAL) [34], which can decrease the computation time of SMLR and KSMLR). The cost function $C = 2^b$ of NLELN is in the range of $= [0, 1, 2, \ldots\ , 10]$ and the kernel function in (12) is used as the Gaussian RBF with $\sigma_{NLELM} = 2^\tau$, $\tau$={-9, -8, …, 0,…8, 9}, and the parameters is set as b=9, $\tau$ =-1. It should be noting that the parameters setting of NLELM is to choose the best parameters in our experiments. For LELM, hidden node L in (3) is a very important parameter and we will evaluate the impact in the next subsection. The parameter $\mu$ in (15) is a tunable parameter controlling the degree of smoothness, which is set as $\mu = 20$ for Indian Pines and Pavia University. We will further evaluate the impact on the proposed approach in the next subsection. It should be noting that the output of LELM and NLELM are probability output. All the experiments are conducted in MATLAB R2016b on a computer with 3.50GHz CPU and 32.0G RAM.

### 3.2. Impact of parameters L and $\mu$

In this subsection, we will evaluate the impact of the hidden neurons of LELM, L, and the smoothness parameter, μ, using the Indian Pines and Pavia University datasets. Table 1 displays the numbers of training samples and test samples.

Figure 3 shows the OA, AA and kappa statistic results as a function of variable L with the training samples of 1043 and 3921 in the Indian Pines and Pavia University, respectively (about 9% and 10% of the total samples, respectively). The training samples are randomly selected from each class in each Monte Carlo Run. From Figure 3 (a) and (b), we can see that the classification accuracies of LELM indeed depend on the hidden neurons, so we should choose the best hidden neurons for LELM in order to improve the classification performance in the sequential spatial information classification. We can see that the best hidden neurons of LELM for Indian Pines is about 450 and the best hidden neurons of LELM for Pavia University is about 1050. Therefore, we will set the hidden neurons as 450 for Indian Pines and 1050 for Pavia University.

Figure 4 (a) and (b) show the OA, AA and kappa statistic as a function of variable $\mu$, we can see that the performance of the proposed framework depend on the smoothness parameter. However, the classification performance is very stable and keeps the high classification accuracy as μ is increasing. This also demonstrates the proposed framework is very robust.

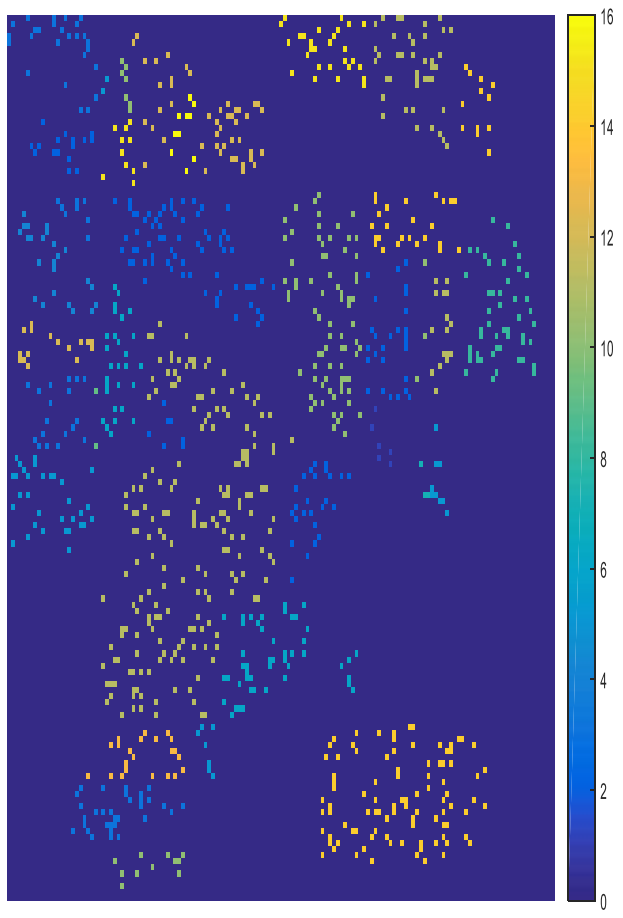

**Figure 5.** AVIRIS Indian Pines training maps.

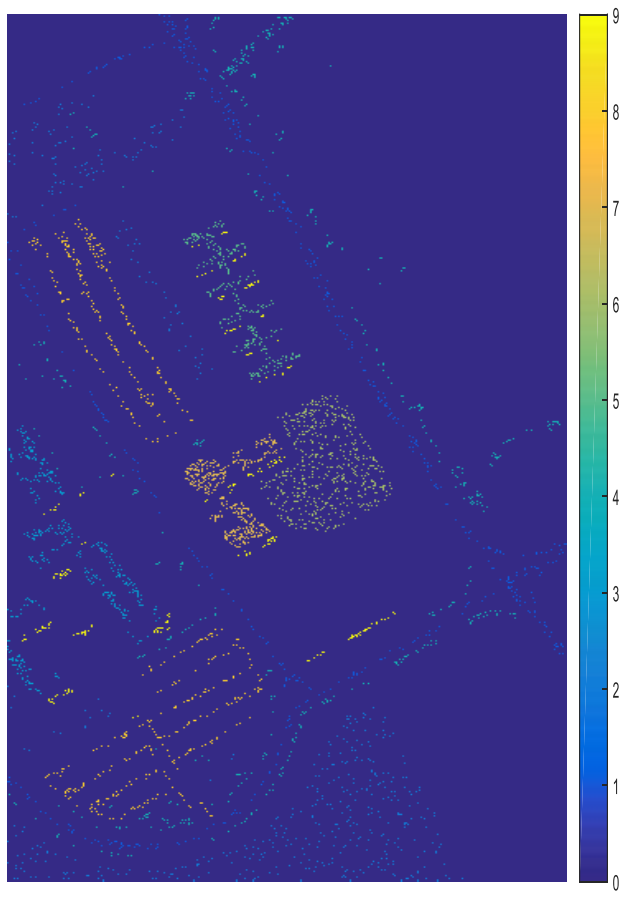

**Figure 6.** Pavia University training maps.

*3.3. The Experiment Resutls and Analysis*

In this subsection, we will evaluate the HSI classification accuracy of the proposed method in the two HSI datasets by comparing with other state-of-the-art methods, includingd the sparse multinomial logistic regression (SMLR) , kernel sparse multinomial logistic regression (KSMLR) [3], nonlinear ELM (NLELM), linear ELM (LELM) [13], SMLR+LBP, KSMLR+LBP, and NLELM+LBP. For the normalization, we use Max method as equation (1) for all the algorithms. Table 1 shows the numbers of training sample and testing sample of Indian Pines and Pavia University.

For illustration, Figure 5 shows the training samples of the Indian Pines data. Figure 7 (a)-(h) shows the classification results obtained by different methods for the Indian Pines data. Moreover, Table 2 shows all the comparable results of different classifiers. From Table 2, it is obvious that the classifiers with spatial information (The proposed method, NLELM-LBP, SMLR-LBP, KSMLR-LBP) have shown a clear advantage over pixel-only counterpart. NLELM obtains the best pixel-only classification results, but the results of NLELM-LBP are not good. This validates that the nonlinear transform will disturb the original salient feature of the original pixels. The reason of the bad results of SMLR is may due to SMLR needs to iterate and the outputs of SMLR will also disturb the original salient feature of pixels. KSMLR-LBP achieves slightly higher result than SMLR-LBP.

The kernel operation is better than non-kernel operation with the pixel-only classifier. Nevertheless, the result of KSMLR-LBP is still lower than the proposed method. Our proposed spectral-spatial method based on LELM and ILBP achieves the best recognition results, comparing with LELM, NLELM, SMLR, KSMLR, NLELM-LBE, SMLR-LBP, KSMLR-LPB. This is due to the usage of the linear transform to keep the original salient features of pixel, and the ILBP to extract the spatial features.

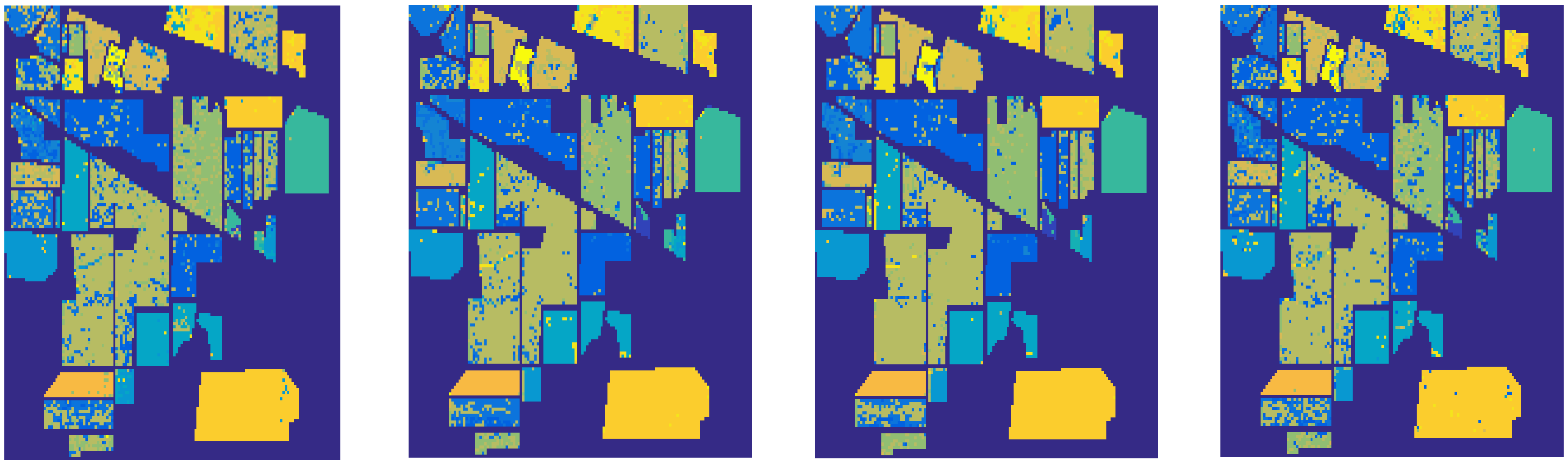

(a) SMLR (b) KSMLR (c) NLELM (d) LELM

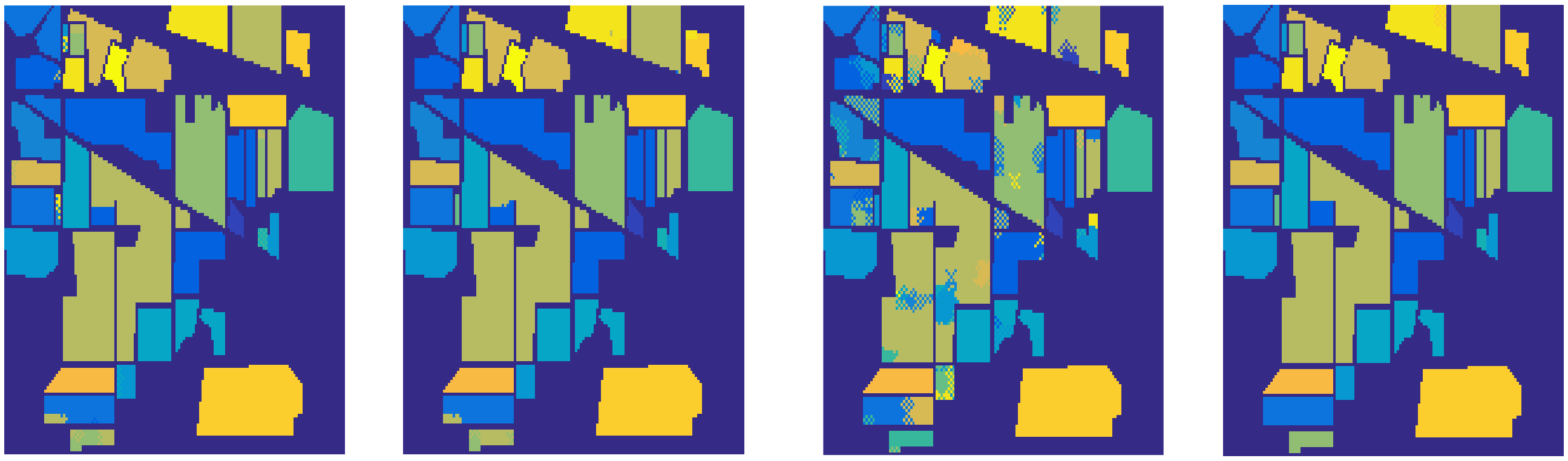

(e) SMLR-LBP (f) KSMLR-LBP (g) NLELM-LBP (h) Proposed method

**Figure 7.** The overall accuracy of Indian Pines image: (a) SMLR (OA=75.76%); (b) KSMLR (OA=84.34%); (c) NLELM (OA=86.93%); (d) LELM (OA=79.43%); (e) SMLR-LBP (OA=98.26%); (f) KSMLR-LBP (OA=99.05%); (g) NLELM-LBP (OA=87.95%); (h) Proposed method (OA=99.75%).

Figure 6 shows the training samples of Pavia University, and Figure 8 shows the classification results of Pavia University and the classification details are reported in Table. 3. It can be seen that the proposed framework also achieves the highest accuracy among all the methods.

**Table 1.** The training sample and test samples of Indian Pines and Pavia University

| **Indian Pines** | | | | | | **Pavia University** | | |
|---|---|---|---|---|---|---|---|---|
| Class | Train | Test | Class | Train | Test | Class | Train | Test |
| Alfalfa | 6 | 54 | Oats | 2 | 20 | Asphalt | 548 | 6631 |
| Corn-no till | 144 | 1434 | Soybeans-no till | 97 | 968 | Meadows | 548 | 18649 |
| Corn-min till | 84 | 834 | Soybeans-min till | 247 | 2468 | Gravel | 392 | 2099 |
| Corn | 24 | 234 | Soybeans-clean till | 62 | 614 | Trees | 524 | 3064 |
| Grass/pasture | 50 | 497 | Wheat | 22 | 212 | Metal sheets | 265 | 1345 |
| Grass/tree | 75 | 747 | Woods | 130 | 1294 | Bare soil | 532 | 5029 |
| Grass/pasture-mowed | 3 | 26 | Bldg-grass-tree-drives | 38 | 380 | Bitumen | 375 | 1330 |
| Hay-windrowed | 49 | 489 | Stone-steel towers | 10 | 95 | Bricks | 514 | 3682 |
| | | | | | | Shadows | 231 | 947 |
| Total | | | | 1043 | 10366 | Total | 3921 | 42776 |

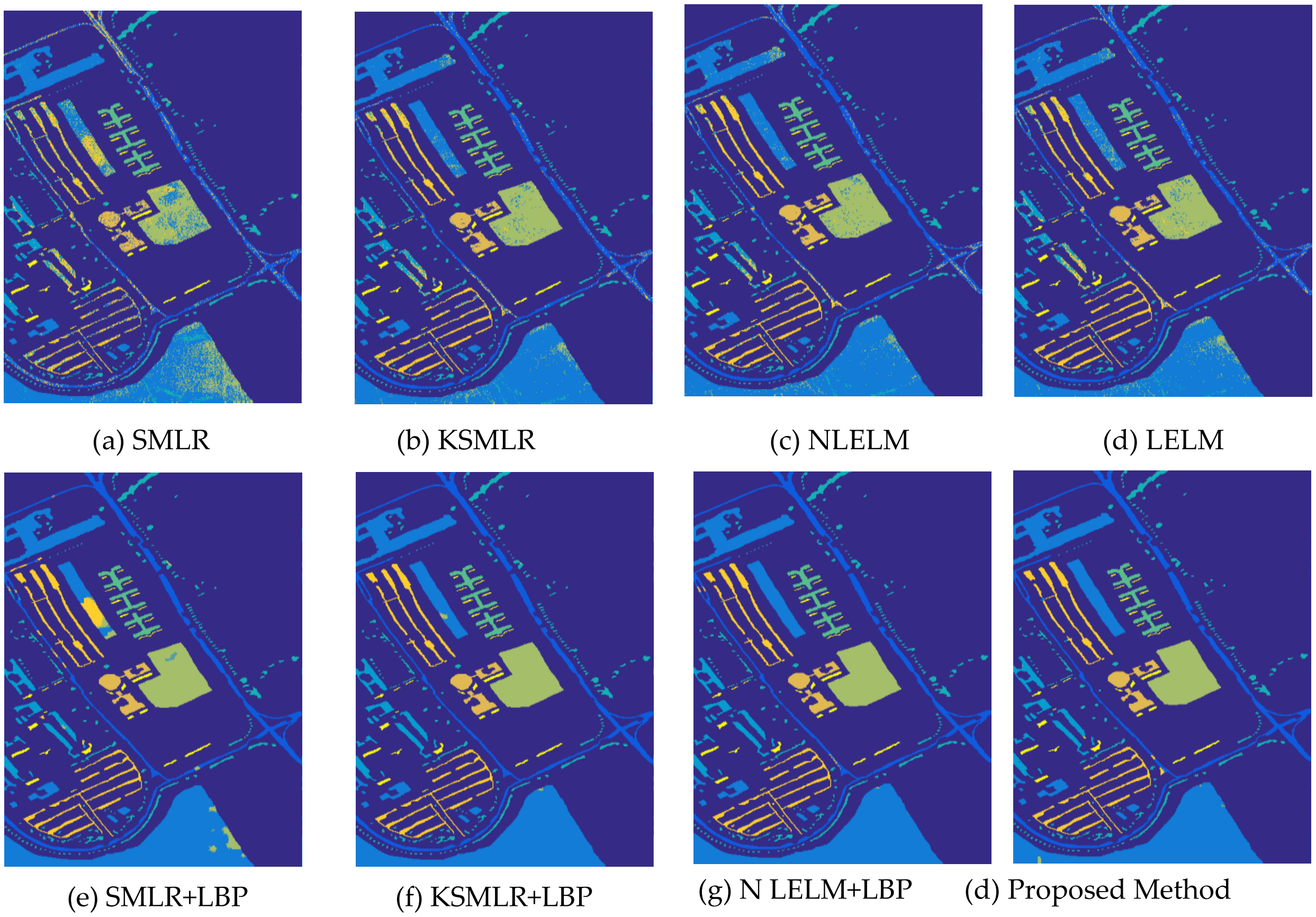

**Figure 8.** The overall accuracy of Pavia University image: (a) SMLR (OA=78.78%); (b) KSMLR (OA=93.00%); (c) NLELM (OA=93.94%); (d) LELM (OA=91.23%); (e) SMLR-LBP (OA=95.68%); (f) KSMLR-LBP (OA=99.42%); (g) NLELM-LBP (OA=99.61%); (h) Proposed method (OA=99.82%).

In the last line of Table 2 and Table 3, we report the average computation time of all the methods on the Indian Pines with 1043 training samples and Pavia University with 3921 training samples. We test for ten Monte Carlo runs, respectively. It is obvious and reasonable that the classifiers with spectral-spatial information cost more time than the pixel-only counterpart. From the last line of Table 2, we can also see that the proposed method has almost the similar computation time as SMLR+LBP for Indian Pines. However, the proposed method achieves higher classification accuracy than SMLR-LBP. The proposed method achieves higher classification accuracy than NLELM-LBP and KSMLR-LBP with much less computation time. From the last line of Table 3, we can get the same conclusion for the Pavia University database. To sum up, the proposed method has achieved higher accuracy than KSMLR-LBP, NLELM-LPB with much less computation time. It is obvious that the proposed LELM-LBP keeps the salient features of HSI very well, so it can obtain higher accuracy than other spectral-spatial method with high computational efficiency.

**Table 2.** Indian Pines: overall, average, and individual class accuracy (in percent) and k statistic of different classification methods with 10% training samples. The best accuracy in each row is show in bold.

| **Class** | **SMLR** | **KSMLR** | **LELM** | **NLELM** | **SMLR-LBP** | **KSMLR-LBP** | **NLELM-LBP** | **PROPOSED METHOD** |
|---|---|---|---|---|---|---|---|---|
| Alfalfa | 30.52 | 74.26 | 35.37 | 71.11 | 97.78 | 100 | 90.37 | **100.00** |
| Corn-no till | 75.87 | 82.49 | 79.27 | 85.82 | 99.02 | 99.40 | 85.68 | **99.68** |
| Corn-min till | 51.35 | 70.86 | 58.26 | 72.58 | 92.55 | 97.35 | 68.79 | **99.22** |
| Corn | 37.35 | 68.68 | 43.29 | 69.10 | 99.27 | 95.00 | 77.44 | **100.00** |
| Grass/pasture | 86.82 | 89.46 | 89.76 | 93.64 | 97.36 | 98.23 | 93.64 | **99.28** |
| Grass/tree | 94.28 | 96.37 | 96.32 | 97.39 | **100.00** | **100.00** | 95.70 | **100.00** |
| Grass/pasture-mowed | 6.92 | 45.00 | 11.54 | 70.38 | 71.92 | 91.54 | 45.00 | **95.38** |
| Hay-windrowed | 99.37 | 98.51 | 99.57 | 99.04 | **100.00** | 100 | 98.73 | **100.00** |
| Oats | 5 | 38.50 | 11.50 | 63.50 | 16.50 | 100 | 48.00 | **100.00** |
| Soybeans-no till | 61.03 | 74.91 | 66.69 | 80.79 | 96.27 | 96.34 | 80.74 | **99.23** |
| Soybeans-min till | 74.46 | 84.51 | 80.23 | 87.66 | **99.96** | 99.91 | 90.41 | 99.93 |
| Soybeans-clean till | 68.96 | 82.20 | 72.98 | 84.98 | 98.50 | **100** | 82.85 | **100.00** |
| Wheat | 96.75 | 99.15 | 99.39 | 98.96 | **100.00** | **100** | 98.77 | **100.00** |
| Woods | 95.04 | 95.20 | 95.65 | 96.51 | **100.00** | 99.69 | 97.26 | **100.00** |
| Bldg-grass-tree-drives | 67.13 | 73.05 | 64.08 | 70.45 | 95.47 | 99.50 | 83.53 | **99.89** |
| Stone-steel towers | 69.26 | 70.32 | 70.42 | 77.05 | 99.58 | 98.63 | 98.63 | **99.89** |
| OA | 75.76 | 84.34 | 79.43 | 86.93 | 98.26 | 99.05 | 87.95 | **99.75** |
| AA | 63.66 | 77.72 | 67.15 | 82.44 | 91.51 | 98.47 | 83.47 | **99.53** |
| k | 72.22 | 82.09 | 76.38 | 85.06 | 98.02 | 98.92 | 86.36 | **99.72** |
| Execution Time (seconds) | 0.02 | 0.41 | 0.19 | 0.31 | **38.74** | 40.70 | 39.59 | 38.95 |

**Table 3.** Pavia University: overall, average, and individual class accuracy (in percent) and k statistic of different classification methods with 10% training samples. The best accuracy in each row is show in bold.

| **Class** | **SMLR** | **KSMLR** | **LELM** | **NLELM** | **SMLR-LBP** | **KSMLR-LBP** | **NLELM-LBP** | **PROPOSED METHOD** |
|---|---|---|---|---|---|---|---|---|
| Asphalt | 72.27 | 89.43 | 85.27 | 88.82 | 98.62 | **99.63** | 99.49 | **99.63** |
| Meadows | 79.08 | 94.16 | 92.17 | 94.61 | 93.70 | 99.34 | 99.88 | **99.83** |
| Gravel | 71.99 | 85.08 | 78.06 | 87.41 | 99.14 | 99.64 | 99.92 | **99.83** |
| Trees | 94.90 | 97.92 | 97.38 | 98.16 | 99.27 | **99.86** | 98.54 | 99.64 |
| Metal sheets | 99.58 | 99.34 | 98.85 | 99.39 | **100.00** | **100.00** | 100.00 | **100.00** |
| Bare soil | 74.26 | 94.77 | 93.90 | 95.43 | 99.93 | **100.00** | 100.00 | **100.00** |
| Bitumen | 78.66 | 93.82 | 93.69 | 95.34 | **100.00** | **100.00** | 100.00 | **100.00** |
| Bricks | 73.37 | 87.52 | 90.05 | 90.94 | 99.93 | 99.63 | 99.85 | **100.00** |
| Shadows | 96.88 | 99.61 | 99.70 | 99.97 | **99.89** | 99.87 | 94.14 | **99.89** |
| OA | 78.78 | 93.00 | 91.23 | 93.94 | 96.93 | 99.59 | 99.62 | **99.83** |
| AA | 82.33 | 93.49 | 92.12 | 94.56 | 98.94 | 99.77 | 99.09 | **99.87** |
| k | 72.73 | 90.82 | 88.54 | 92.04 | 95.98 | 99.46 | 99.49 | **99.78** |
| Execution Time (seconds) | 0.19 | 4.40 | 0.48 | 3.83 | **1193.7** | 1237.1 | 5288.6 | 1201.2 |

## 4. Conclusions

In this work, we had proposed a new framework for HSI classification using spectral-spatial information with LELM and LBP. The LELM method is used to learn a spectral classifier for the original HSI data and keep the salient features of HSI. The spatial information is modeled based on LBP in order to improve the classification accuracy of HSI. The proposed method keeps the salient feature of HSI for the spatial-based classification. Experiment results show the superiority of the proposed method.

In the future work, we will focus on learning the dictionary of each class in the spectral domain for LELM in order to further improve the classification of LELM. In order to improve the classification results furtherly, we will resort to Spatial Filtering [35]. Moreover, we will also decrease the time-consuming by resort the extended multi-attribute profiles (EMAPs) [36] method.

**Acknowledgments:** This work is supported by the National Nature Science Foundation of China (no. 61471132, 61372173), the Training program for outstanding young teachers in higher education institutions of Guangdong Province (no. YQ2015057)